\pdfoutput=1
\documentclass[11pt]{article}

\usepackage[preprint]{acl}
\usepackage{enumitem} % Add this in your preamble

\usepackage{times}
\usepackage{latexsym}

\usepackage[T1]{fontenc}
\usepackage[utf8]{inputenc}

\usepackage{microtype}

\usepackage{inconsolata}

\usepackage{graphicx}

\usepackage{float}
\usepackage{booktabs}
\usepackage{array}
\newcolumntype{P}[1]{>{\centering\arraybackslash}m{#1}}

\title{Can Agents Judge Systematic Reviews Like Humans? Evaluating SLRs with LLM-based Multi-Agent System}

\author{
Abdullah Mushtaq \\
Department of Computer Science \\
Information Technology University\\
\small{bscs20078@itu.edu.pk} 
\And
Muhammad Rafay Naeem \\
Department of Computer Science \\
Information Technology University \\
\small{bscs20004@itu.edu.pk} 
\AND
Ibrahim Ghaznavi \\
Department of Computer Science \\
Information Technology University \\
\small{ibrahim.ghaznavi@itu.edu.pk} 
\And
Alaa Abd-alrazaq \\
AI Center for Precision Health \\
Weill Cornell Medicine–Qatar \\
\small{aaa4027@qatar-med.cornell.edu} 
\AND
Aliya Tabassum \\
Computer Science and \\Engineering Department\\
Qatar University \\
\small{aliyatabassum.jntu@gmail.com} 
\And
Junaid Qadir \\
Computer Science and \\Engineering Department\\
Qatar University \\
\small{jqadir@qu.edu.qa} 
}

\begin{document}
\maketitle
\vspace{5cm}
\begin{abstract}
Systematic Literature Reviews (SLRs) are foundational to evidence-based research but remain labor-intensive and prone to inconsistency across disciplines. We present an LLM-based SLR evaluation copilot built on a \textit{Multi-Agent System (MAS)} architecture to assist researchers in assessing the overall quality of the systematic literature reviews. The system automates protocol validation, methodological assessment, and topic relevance checks using a scholarly database. Unlike conventional single-agent methods, our design integrates a specialized agentic approach aligned with PRISMA guidelines to support more structured and interpretable evaluations. We conducted an initial study on five published SLRs from diverse domains, comparing system outputs to expert-annotated PRISMA scores, and observed 84\% agreement. While early results are promising, this work represents a first step toward scalable and accurate NLP-driven systems for interdisciplinary workflows and reveals their capacity for rigorous, domain-agnostic knowledge aggregation to streamline the review process.

\end{abstract}

\section{Introduction}

Systematic Literature Reviews are foundational to evidence-based research, offering a structured, protocol-driven approach for identifying, analyzing, and synthesizing prior work. In contrast to narrative reviews, SLRs follow a predefined methodology to promote transparency, reproducibility, and reduced bias \cite{grant2009typology, booth2021systematic}. They are widely employed to surface research trends, identify knowledge gaps, and establish a grounded basis for future inquiry across disciplines.

\begin{figure}[h]
    \centering
    \includegraphics[width=\linewidth]{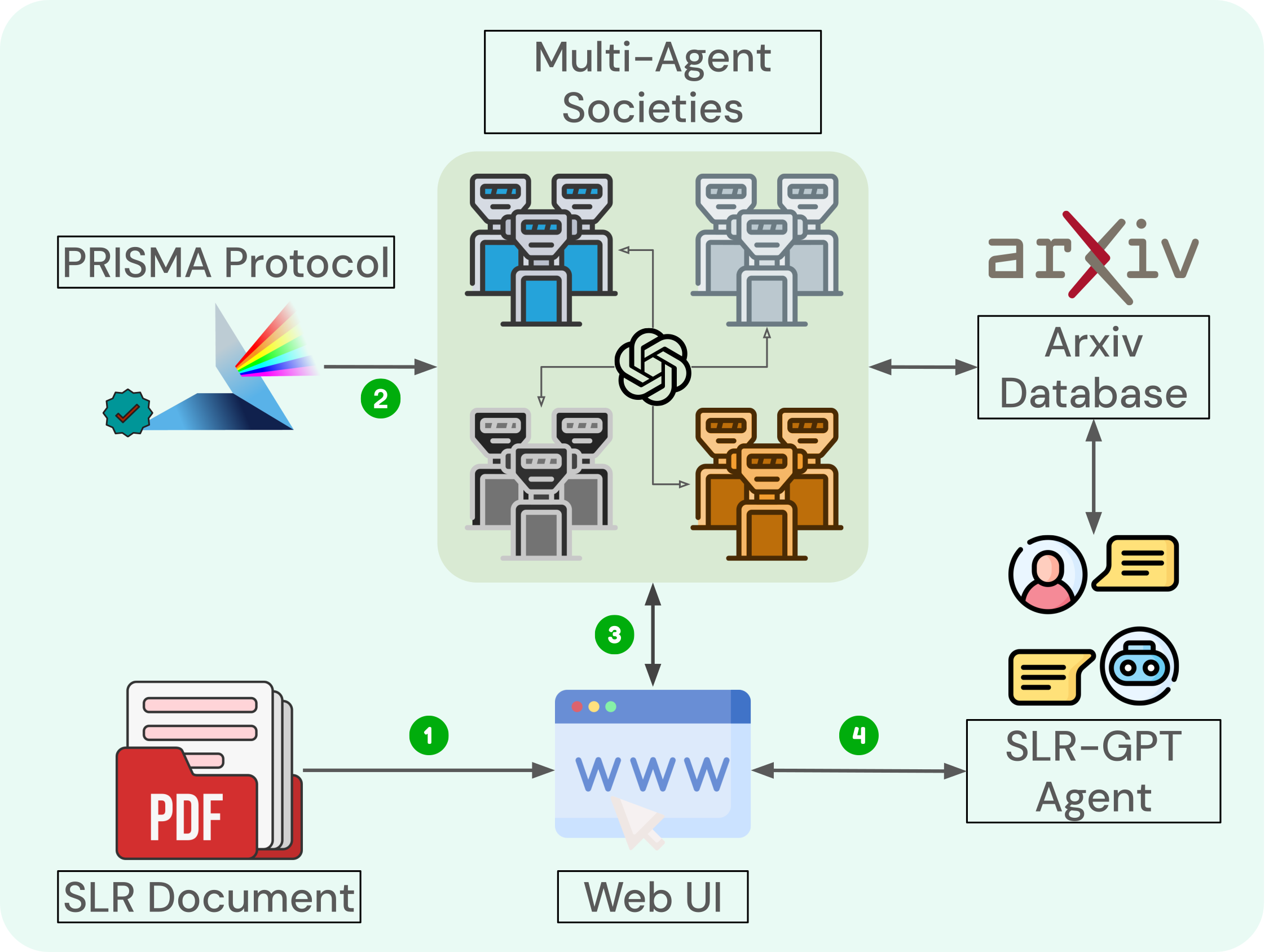}
    \caption{An overview of the proposed multi-agent system LLM-based SLR evaluation framework.}
    \label{fig:Overview}
\end{figure}

However, the exponential growth of scholarly publications, varying in quality and relevance, has made it increasingly challenging to maintain rigor and comprehensiveness in the SLR process. This overload slows down the review pipeline and introduces risks of redundancy and overlooks literature.

To address these challenges, we propose an AI-augmented SLR system that leverages a multi-agent LLM architecture. Inspired by human-centered co-pilot design principles \cite{sellen2024rise}, our system supports the SLR workflow: from protocol critique and methodological assessment to relevance checking, duplication detection, and collaborative drafting. Figure \ref{fig:Overview} provides a high-level overview of the proposed system.

The tool is designed with an interdisciplinary lens, recognizing the cross-domain nature of SLRs, from health sciences to software engineering, while leveraging LLM-based agentic architectures to enhance quality and efficiency using the NLP architecture proposed by \cite{vaswani2017attention}. By combining automation with human oversight, our approach takes an initial step toward improving accessibility and robustness in evidence synthesis. To guide our study, we focus on the practical capabilities and evaluation of the proposed system within the SLR workflow. 

Our research questions (RQs) are as follows:

\begin{enumerate}[label=\textbf{RQ\arabic*}:]
    \item How can a multi-agent LLM system support the protocol validation and compliance steps of the SLR process?
    
    \item How well does the system's output align with PRISMA standards, based on initial expert evaluations, and does the system offer measurable improvements in efficiency or consistency during SLR evaluation?
\end{enumerate}

\section{Background and Related Work}

SLRs are essential for synthesizing research findings, identifying gaps, and guiding future studies. The PRISMA guidelines \cite{moher2009preferred, page2021prisma} offer a structured approach to ensure transparency and reproducibility. Tools like Covidence \cite{covidence} and Rayyan \cite{rayyan} assist in screening and data extraction but rely heavily on manual effort, making them prone to fatigue, selection bias, and inconsistency.

Recent work explores leveraging LLMs for automating SLR stages such as study identification, summarization, and quality assessment \cite{susnjak2023prisma, ge2024ai, smith2023sentiment, jones2022automated}. State-of-the-art LLMs like GPT-4, Claude 3.7, Llama, and Gemini 2.5 \cite{GPT4_1, Gemini, sonnet3.7, Llama4} demonstrate impressive few-shot learning capabilities \cite{LLMs_FewShotLearners}, making them suitable for structured tasks with minimal supervision.

To enhance performance on complex, multi-step tasks, MAS have emerged as powerful frameworks capable of decomposing problems, enabling cooperative reasoning, and outperforming single-agent models on structured benchmarks \cite{mast2025failures, ma_survey_2025, wang2024challenges}. Recent MAS-driven tools illustrate this shift: Google's AI Co‑Scientist \cite{google2025aicoscientist} leverages Gemini-based agents to generate hypotheses and propose experiments; OpenAI's Deep Research \cite{openai2025deepresearch} conducts autonomous literature synthesis via web search; and SciSpace \cite{scispace2025} offers interactive document parsing and drafting.

While existing tools support general scientific exploration and offer basic workflow automation, they often lack the methodological rigor required for systematic reviews. Similarly, recent studies underscore the limitations of core monolithic LLMs in structured tasks: \citeauthor{lieberum2025scoping} reviewed 37 GPT-based SR prototypes and found them largely unvalidated; \citeauthor{penzo2024multiparty} demonstrated that LLaMA2-13B exhibits prompt and structure sensitivity; and \citeauthor{wei2022chain, wang2023self} reported only modest gains, ranging from 3.9\% to 17.9\%, from techniques like chain-of-thought and self-consistency across various benchmarks. These findings suggest inherent performance ceilings in end-to-end LLM pipelines for SRs. To address the limitations posed by monolithic LLMs, we introduce a modular, multi-agent LLM system explicitly aligned with PRISMA guidelines, where each checklist item is handled by a specialized agent under expert oversight. Early results indicate improved consistency and reliability. Our open-source implementation\footnote{GitHub link will be added here upon publication} promotes transparency, supports scalable, high-quality reviews, and provides a foundation for robust, end-to-end SLR support.

\begin{figure*}[!t]
  \centering
  \includegraphics[width=\textwidth]{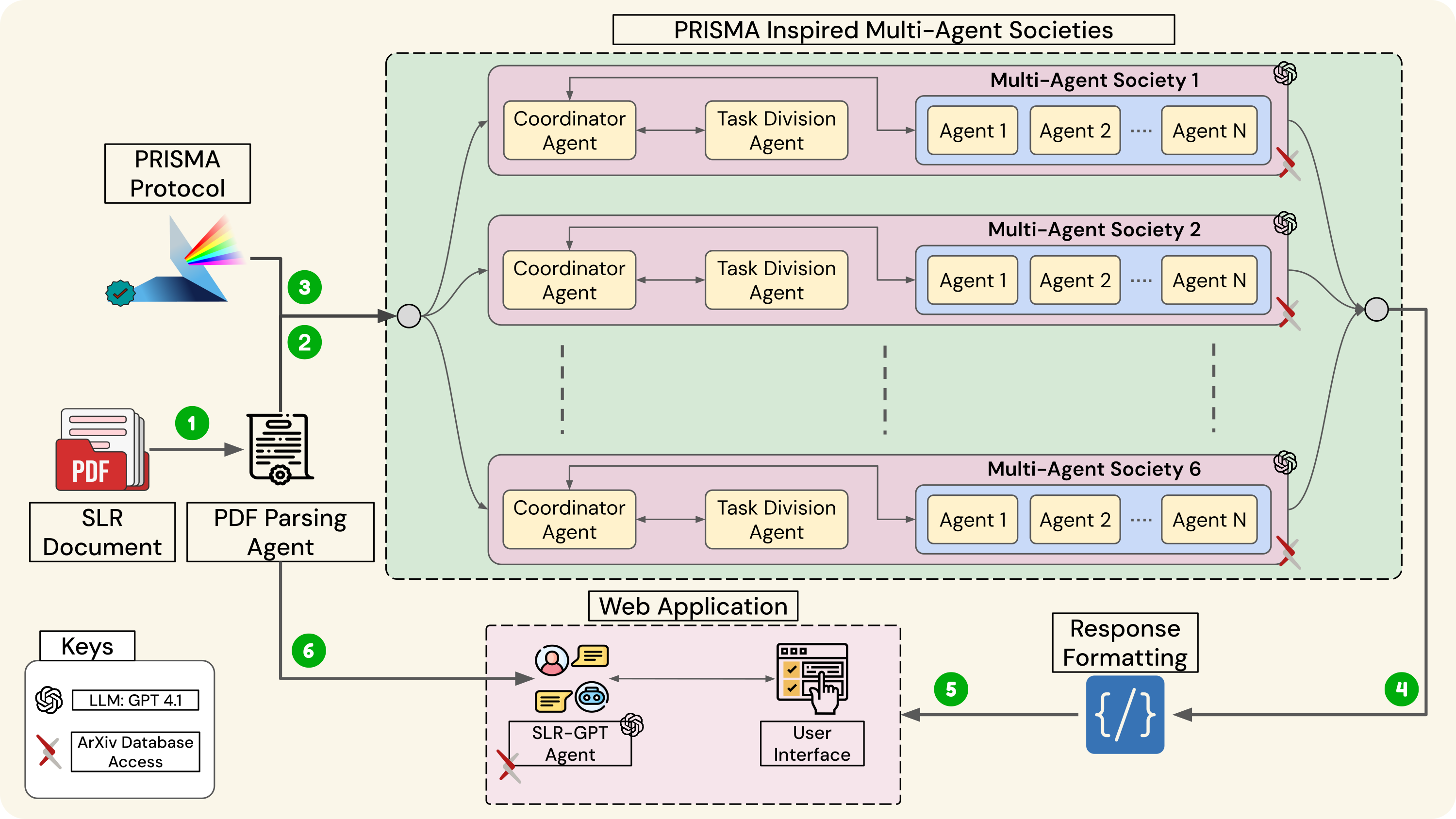}
  \caption{Architecture of the MAS‑LLM SLR evaluation framework. The user-uploaded SLR document is processed by multiple agentic societies designed to evaluate it according to PRISMA guidelines. The results are displayed on the web user interface, which also supports follow-up questions and interactive conversations.}
  \label{fig:Detailed_System_Design}
\end{figure*}

\section{Methodology \& System Design} \label{sec:methodology}

\subsection{Architecture}

Our MAS‑LLM SLR Evaluation Framework consists of 27 specialized agents organized into six PRISMA‑aligned societies (see Table~\ref{tab:societies}) along with two utility agents (PDF Parsing and Follow‑up Conversation). This structure (number of agents and division in societies) mirrors the PRISMA checklist which clearly elicits the checklist items for each part of the systematic reviews \cite{page2021prisma, susnjak2023prisma}, with one agent per checklist item. Sections like \textbf{Methods} have more agents (11) due to their detailed protocol requirements (greater number of checklist items), while sections like \textbf{Discussion} require only one agent. Early experiments with multi-item agents resulted in unstable behavior---overloaded agents, degraded performance, and unpredictable agent spawning, so we adopted a one-agent-per-item design with one-shot detailed prompting along with examples for robustness and clarity. All agents use GPT‑4.1 \cite{GPT4_1}, a state‑of‑the‑art LLM built for agentic workflows with a 1M token context window. Upon SLR PDF upload, an OCR‑enabled Vision–Language Model \cite{unstructured} converts it into structured text.

A \textbf{Coordinator Agent} and \textbf{Task Agent} decompose the PRISMA checklist into modular evaluation tasks, dispatching them via few-shot prompts to specialized agents. Each agent uses the arXiv Toolkit to retrieve relevant research as needed, assigns a 0–5 score, and provides qualitative feedback. If outputs fall below thresholds, the Coordinator reallocates tasks or spawns new agents. As shown in Figure~\ref{fig:Detailed_System_Design}, agent outputs are synthesized into a unified format, accessible via web-interface and provided to the Follow-up Conversation Agent.

\begin{table}[h]
  \small
  \centering
  \begin{tabular}{|P{1.5cm}|P{4cm}|P{0.9cm}|}
    \hline
    \textbf{Society}       & \textbf{Function}                                                   & \textbf{Agents} \\
    \hline
    Abstract \& Title      & Evaluate title clarity and abstract completeness                     & 2  \\ \hline
    Introduction           & Assess rationale, scope, and objectives                              & 2  \\ \hline
    Methods                & Check eligibility criteria, search strategy, and bias assessment     & 11 \\ \hline
    Results                & Verify result reporting, visualizations, and statistical summaries   & 7  \\ \hline
    Discussion             & Examine interpretation, limitations, and implications                & 1  \\ \hline
    Other Information      & Review registration, funding, conflicts of interest, and data policies & 4  \\ \hline
    Standalone             & PDF parsing and follow‑up dialogue                                   & 2  \\
    \hline
  \end{tabular}
  \caption{Agent societies, their responsibilities, and the number of agents in each society.}
  \label{tab:societies}
\end{table}

The follow‑up conversation agent named  \emph{SLR‑GPT Agent} provides co‑pilot style research support through professional interactions. Using the same arXiv Toolkit available to all agents, it suggests new papers, verifies citations, cross-checks literature results, and recommends editorial refinements to maximize PRISMA compliance. By combining structured evaluation outputs from agents from societies with in‑context retrieval for both the original manuscript and PRISMA protocol, it transforms assessments into actionable guidance for manuscript improvement. Open source implementation \footnote{GitHub link will be added here upon publication} is also shared for reproducibility.

\subsection{Evaluation Methodology} \label{subsec:eval_methodology}
We assess our framework on five published SLRs from diverse fields (Medical, E-commerce, AI, Metaverse, IoT), comparing agent outputs against ratings by three expert SLR reviewers. Both agents and human experts score each PRISMA item on a 0–5 scale (0 = Not Addressed, ..., 5= Thoroughly Addressed), enabling a standardized, ordinal evaluation across sections. Agent prompts incorporate PRISMA guidelines and one‑shot exemplars to standardize evaluation; human reviewers assess the original manuscripts along with PRISMA guidelines. We quantify alignment using Mean Absolute Error (MAE), agreement level using MAE and additional statistical analysis to pinpoint areas where multi‑agent collaboration aligns best with human experts. This benchmark demonstrates the technical viability and interdisciplinary applicability of agentic LLMs in streamlining SLR workflows across diverse scientific domains.

\section{Results}
%In this section, we will answer the RQs mentioned above with our empirical results. 

\subsection{MAE \& Agreement Level}
\begin{figure}[!h]
    \centering
    \includegraphics[width=\linewidth]{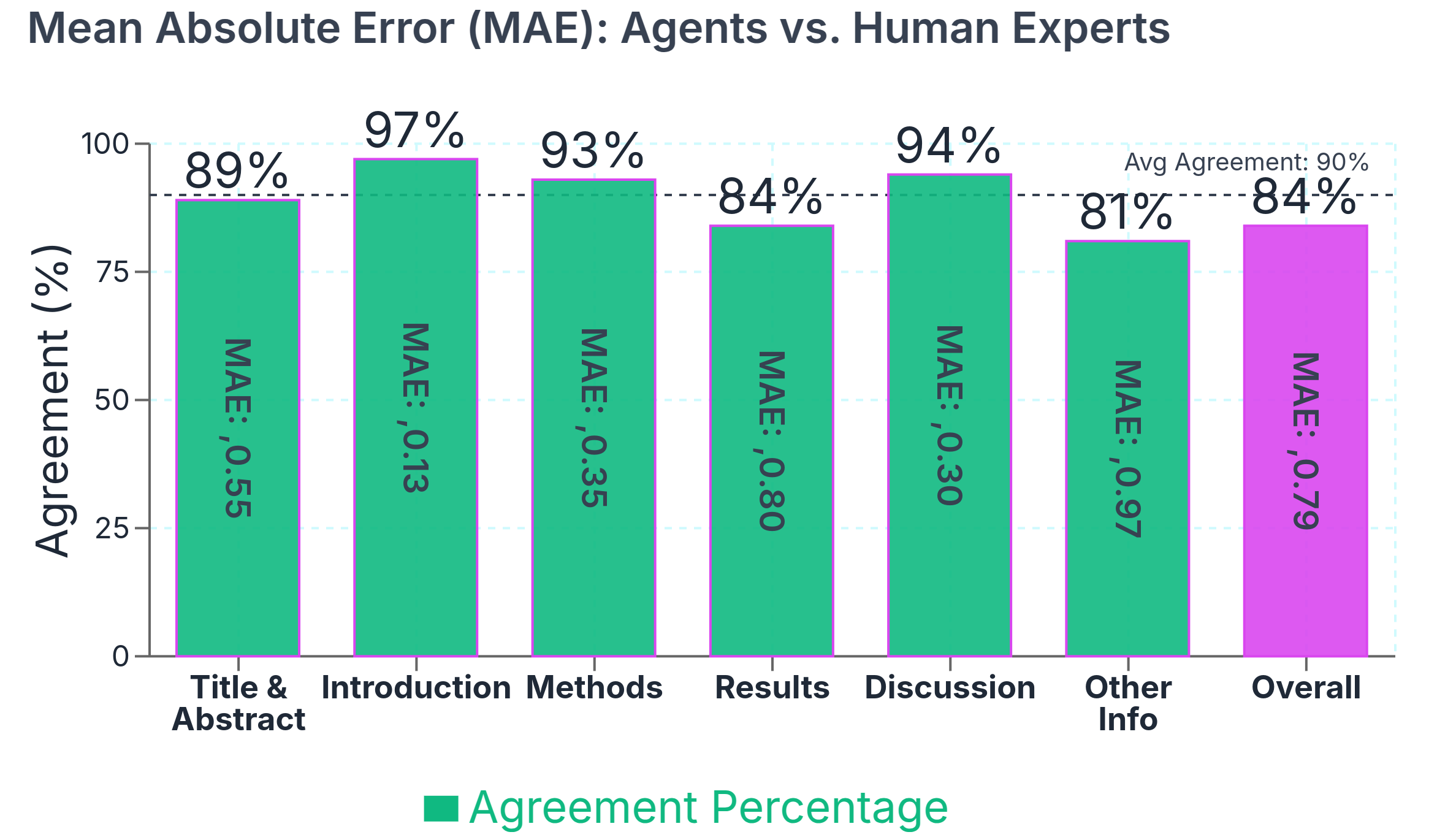}
    \caption{Agreement between agents and humans.
    % computed as $100 - (\mathrm{MAE}/5 \times 100)$.
    Avg.\ agreement is 84\%, with the highest alignment in Introduction (97\%) and lower in Other Information (81\%).}
    \label{fig:agreement}
\end{figure}

We evaluated our system on five published SLRs from diverse domains (Medical, AI, Ecommerce, IoT, Metaverse), comparing its PRISMA‑based scores against expert human reviewers (Section~\ref{subsec:eval_methodology}). Agreement percentages, computed as $100\% - (\mathrm{MAE}/5 \times 100)$, are shown in Figure~\ref{fig:agreement}. 

According to our results, the overall agreement is \textbf{84\%}, with strongest alignment in \emph{Introduction} (97\%), \emph{Discussion} (94\%), and \emph{Methods} (93\%). Alignment dips only slightly to \emph{Results} (84\%) and \emph{Other Information} (81\%). In absolute terms, the highest agreement exceeds the overall agreement by 13 points, while the lowest is just 3 points below, indicating that all section‑level agreements fall within a narrow window around the mean (close alignment overall). These agreement levels demonstrate that our system faithfully reproduces expert judgments on core review components by aligning itself with the PRISMA protocol while highlighting opportunities for future improvement. This alignment with experts also shows that our proposed MAS supports the protocol validation and compliance mainly through its system design and architectural choices mentioned in Section \ref{sec:methodology}.

To address the significant delays in traditional peer review, averaging 15 weeks for the first round, with 10 weeks in medical sciences, 14 in natural sciences, and 17 in social sciences \cite{scirev2014average}, and up to 25 weeks in fields like Economics \cite{huisman2017duration}, we present a automated MAS review system. It analyzes papers in just 15–20 minutes based on length and complexity, offering early-stage insights that can guide and accelerate subsequent human reviews, effectively reducing overall turnaround time.

\subsection{Paper-wise \& Inter-expert Analysis}
Per‐paper analysis (Appendix Fig.\ref{fig:paper_analysis}) shows agent scores consistently track human evaluations across all five SLRs. Expert comparison (Appendix Fig.\ref{fig:expert_comparison}) indicates the highest inter‐rater variability. High inter‐expert agreement---Intraclass Correlation Coefficient = 0.924, Krippendorff's $\alpha = 0.889$, Pearson $\rho = 0.898$ (Appendix Table~\ref{tab:human_agreement}), validates our human benchmark and suggests remaining divergences highlight targets for system refinement.

\section{Future Directions}

To ensure practical utility, we plan to deploy an interactive, browser-based interface that allows users to ask questions, revise summaries, and re-score sections, enabling both quantitative and qualitative evaluation of UI design and agent responsiveness. We will test the system with real-world users, systematic reviewers, and authors to assess its collaborative effectiveness. Structured feedback will be collected using Likert scales to evaluate clarity, usefulness, and trust, following best practices in human–AI interaction \cite{zhao-etal-2024-successfully, rong2022hxai}. This feedback will be integrated into a preference-tuning loop to better align agent behavior with user preferences \cite{shao2025visualization}.

\section{Conclusion}
We introduced a multi-agent system for evaluating Systematic Literature Reviews aligned with the PRISMA protocol. By leveraging specialized agents for protocol validation, topic relevance, structural assessment, and ArXiv integration, the system aims to reduce the manual burden of interdisciplinary SLR evaluation. The built-in SLR-GPT Assistant supports citation checks and editorial feedback. An initial small-scale empirical evaluation on five SLRs from diverse domains showed 84\% agreement with expert SLR judgments. While promising, these results are preliminary. This work offers a first step toward scalable, accurate MAS for SLRs, with future efforts focused on broader evaluations and system refinement with a focus on human-AI collaboration.

\section*{Limitations}

Our MAS-LLM framework shows early promise, but several limitations should be acknowledged. The current evaluation spans five SLRs from distinct domains. In subsequent studies, we are looking towards increasing the number of papers and hence the credibility of our results. Agent performance is bounded by the capabilities of current LLMs, which may overlook fine-grained domain knowledge in technical contexts. The system's integration with arXiv enhances open-access coverage but excludes other key databases like PubMed or Scopus, creating potential gaps. Moreover, the system currently supports only evaluation, not real-time drafting or collaboration. Nonetheless, this study demonstrates the feasibility of structured, agentic LLM support for SLRs and lays the groundwork for more scalable and interactive systems in future work.

\bibliography{bib}

\begin{thebibliography}{33}
\providecommand{\natexlab}[1]{#1}

\bibitem[{Anthropic-AI(2025)}]{sonnet3.7}
Anthropic-AI. 2025.
\newblock {Claude 3.7 Sonnet and Claude Code}.
\newblock Technical report, Anthropic AI.
\newblock Retrieved from \url{https://www.anthropic.com/news/claude-3-7-sonnet}.

\bibitem[{Booth et~al.(2021)Booth, James, Clowes, Sutton et~al.}]{booth2021systematic}
Andrew Booth, Martyn-St James, Mark Clowes, Anthea Sutton, and 1 others. 2021.
\newblock Systematic approaches to a successful literature review.

\bibitem[{Brown et~al.(2020)Brown, Mann, Ryder, Subbiah, Kaplan, Dhariwal, Neelakantan, Shyam, Sastry, Askell et~al.}]{LLMs_FewShotLearners}
Tom Brown, Benjamin Mann, Nick Ryder, Melanie Subbiah, Jared~D Kaplan, Prafulla Dhariwal, Arvind Neelakantan, Pranav Shyam, Girish Sastry, Amanda Askell, and 1 others. 2020.
\newblock Language models are few-shot learners.
\newblock \emph{Advances in neural information processing systems}, 33:1877--1901.

\bibitem[{{Covidence Systematic Review Software}(2024)}]{covidence}
{Covidence Systematic Review Software}. 2024.
\newblock Covidence: Streamlining systematic review workflow.
\newblock Online resource.
\newblock Available at \url{https://www.covidence.org/}.

\bibitem[{DeepMind(2025)}]{Gemini}
DeepMind. 2025.
\newblock Gemini: Revolutionizing {AI} with multimodal capabilities.
\newblock Technical report, DeepMind.
\newblock Retrieved from \url{https://deepmind.google/technologies/gemini/}.

\bibitem[{Ge and Others(2024)}]{ge2024ai}
X~Ge and Others. 2024.
\newblock {AI}-driven systematic reviews in health research.
\newblock \emph{Medical Informatics Journal}.

\bibitem[{Gottweis et~al.(2025)Gottweis, Weng, Daryin, Tu, Palepu, Sirkovic, Myaskovsky, Weissenberger, Rong, Tanno et~al.}]{google2025aicoscientist}
Juraj Gottweis, Wei-Hung Weng, Alexander Daryin, Tao Tu, Anil Palepu, Petar Sirkovic, Artiom Myaskovsky, Felix Weissenberger, Keran Rong, Ryutaro Tanno, and 1 others. 2025.
\newblock Towards an {AI} co-scientist.
\newblock \emph{arXiv preprint arXiv:2502.18864}.

\bibitem[{Grant and Booth(2009)}]{grant2009typology}
Maria~J Grant and Andrew Booth. 2009.
\newblock A typology of reviews: an analysis of 14 review types and associated methodologies.
\newblock \emph{Health information \& libraries journal}, 26(2):91--108.

\bibitem[{Huisman and Smits(2017)}]{huisman2017duration}
Jeroen Huisman and Jeroen Smits. 2017.
\newblock Duration and quality of the peer review process: the author's perspective.
\newblock \emph{Scientometrics}, 113(1):633--650.

\bibitem[{Jones and Others(2022)}]{jones2022automated}
L~Jones and Others. 2022.
\newblock Automating sentiment analysis in systematic reviews.
\newblock \emph{AI Review Journal}.

\bibitem[{Lieberum et~al.(2025)Lieberum, T{\"o}ws, Metzendorf, Heilmeyer, Siemens, Haverkamp, B{\"o}hringer, Meerpohl, and Eisele-Metzger}]{lieberum2025scoping}
Judith-Lisa Lieberum, Markus T{\"o}ws, Maria-Inti Metzendorf, Felix Heilmeyer, Waldemar Siemens, Christian Haverkamp, Daniel B{\"o}hringer, Joerg~J. Meerpohl, and Angelika Eisele-Metzger. 2025.
\newblock \href {https://doi.org/10.1016/j.jclinepi.2025.111746} {Large language models for conducting systematic reviews: on the rise, but not yet ready for use---a scoping review}.
\newblock \emph{Journal of Clinical Epidemiology}, 181:111746.

\bibitem[{Meta-AI(2025)}]{Llama4}
Meta-AI. 2025.
\newblock The llama 4 herd: The beginning of a new era of natively multimodal {AI} innovation.
\newblock Technical report, Meta AI.
\newblock Retrieved from \url{https://ai.meta.com/blog/llama-4-multimodal-intelligence/}.

\bibitem[{Moher et~al.(2009)Moher, Liberati, Tetzlaff, and Altman}]{moher2009preferred}
David Moher, Alessandro Liberati, Jennifer Tetzlaff, and Douglas~G Altman. 2009.
\newblock Preferred reporting items for systematic reviews and meta-analyses: The {PRISMA} statement.
\newblock \emph{BMJ}, 339:b2535.

\bibitem[{OpenAI(2025{\natexlab{a}})}]{openai2025deepresearch}
OpenAI. 2025{\natexlab{a}}.
\newblock Introducing deep research.
\newblock \url{https://openai.com/index/introducing-deep-research/}.
\newblock Accessed: 2025-05-06.

\bibitem[{OpenAI(2025{\natexlab{b}})}]{GPT4_1}
OpenAI. 2025{\natexlab{b}}.
\newblock \href {https://openai.com/index/gpt-4-1/} {Introducing {GPT-4.1} in the {API}}.

\bibitem[{Ouzzani et~al.(2016)Ouzzani, Hammady, Fedorowicz, and Elmagarmid}]{rayyan}
Mourad Ouzzani, Hossam Hammady, Zbys Fedorowicz, and Ahmed Elmagarmid. 2016.
\newblock \href {https://doi.org/10.1186/s13643-016-0384-4} {Rayyan---a web and mobile app for systematic reviews}.
\newblock \emph{Systematic Reviews}, 5:210.

\bibitem[{Page et~al.(2021)Page, McKenzie, Bossuyt, Boutron, Hoffmann, Mulrow, Shamseer, Tetzlaff, Akl, Brennan et~al.}]{page2021prisma}
Matthew~J Page, Joanne~E McKenzie, Patrick~M Bossuyt, Isabelle Boutron, Tammy~C Hoffmann, Cynthia~D Mulrow, Larissa Shamseer, Jennifer~M Tetzlaff, Elie~A Akl, Sue~E Brennan, and 1 others. 2021.
\newblock The {PRISMA} 2020 statement: an updated guideline for reporting systematic reviews.
\newblock \emph{BMJ}, 372:n71.

\bibitem[{Park et~al.(2025)Park, Roesner, Kohno, and Weld}]{mast2025failures}
Sooyoung Park, Franziska Roesner, Tadayoshi Kohno, and Daniel Weld. 2025.
\newblock Why do multi-agent {LLM} systems fail?
\newblock \emph{arXiv preprint arXiv:2503.13657}.

\bibitem[{Penzo et~al.(2024)Penzo, Sajedinia, Lepri, Tonelli, and Guerini}]{penzo2024multiparty}
Nicol{\`o} Penzo, Maryam Sajedinia, Bruno Lepri, Sara Tonelli, and Marco Guerini. 2024.
\newblock Do {LLMs} suffer from multi-party hangover? a diagnostic approach to addressee recognition and response selection in conversations.
\newblock In \emph{Proceedings of the 2024 Conference on Empirical Methods in Natural Language Processing}, pages 11210--11233. Association for Computational Linguistics.

\bibitem[{Rong et~al.(2022)Rong, Leemann, Nguyen, Fiedler, Qian, Unhelkar, Seidel, Kasneci, and Kasneci}]{rong2022hxai}
Yao Rong, Tobias Leemann, Thai-Trang Nguyen, Lisa Fiedler, Peizhu Qian, Vaibhav Unhelkar, Tina Seidel, Gjergji Kasneci, and Enkelejda Kasneci. 2022.
\newblock Towards human-centered explainable ai: A survey of user studies for model explanations.
\newblock \emph{arXiv preprint arXiv:2210.11584}.

\bibitem[{{SciRev}(2014)}]{scirev2014average}
{SciRev}. 2014.
\newblock \href {https://scirev.org/news/average-duration-first-review-round-15-weeks/} {Average duration first review round 15 weeks}.
\newblock Accessed: 2025-05-18.

\bibitem[{SciSpace(2025)}]{scispace2025}
SciSpace. 2025.
\newblock Scispace: {AI} chat for scientific pdfs.
\newblock \url{https://typeset.io/}.
\newblock Accessed: 2025-05-06.

\bibitem[{Sellen and Horvitz(2024)}]{sellen2024rise}
Abigail Sellen and Eric Horvitz. 2024.
\newblock The rise of the {AI} co-pilot: Lessons for design from aviation and beyond.
\newblock \emph{Communications of the ACM}, 67(7):18--23.

\bibitem[{Shao et~al.(2025)Shao, Shan, He, Yao, Wang, Zhang, Zhang, and Chen}]{shao2025visualization}
Zekai Shao, Yi~Shan, Yixuan He, Yuxuan Yao, Junhong Wang, Xiaolong Zhang, Yu~Zhang, and Siming Chen. 2025.
\newblock Do language model agents align with humans in rating visualizations? an empirical study.
\newblock \emph{arXiv preprint arXiv:2505.06702}.

\bibitem[{Smith and Others(2023)}]{smith2023sentiment}
J~Smith and Others. 2023.
\newblock Sentiment analysis in systematic literature reviews.
\newblock \emph{Computational Linguistics}.

\bibitem[{Susnjak(2023)}]{susnjak2023prisma}
Teo Susnjak. 2023.
\newblock {PRISMA-DFLLM}: An extension of {PRISMA} for systematic literature reviews using domain-specific finetuned large language models.
\newblock \emph{arXiv preprint arXiv:2306.14905}.

\bibitem[{{Unstructured Technologies, Inc.}(2025)}]{unstructured}
{Unstructured Technologies, Inc.} 2025.
\newblock \href {https://unstructured.io/} {Unstructured - your {GenAI} has a data problem}.
\newblock Accessed: February 27, 2025.

\bibitem[{Vaswani et~al.(2017)Vaswani, Shazeer, Parmar, Uszkoreit, Jones, Gomez, Kaiser, and Polosukhin}]{vaswani2017attention}
Ashish Vaswani, Noam Shazeer, Niki Parmar, Jakob Uszkoreit, Llion Jones, Aidan~N Gomez, {\L}ukasz Kaiser, and Illia Polosukhin. 2017.
\newblock Attention is all you need.
\newblock \emph{Advances in neural information processing systems}, 30.

\bibitem[{Wang et~al.(2023)Wang, Zhou, Schuurmans et~al.}]{wang2023self}
Xuezhi Wang, Yixin Zhou, Dale Schuurmans, and 1 others. 2023.
\newblock Self-consistency improves chain of thought reasoning in language models.
\newblock In \emph{Proceedings of the 2023 EMNLP}, pages 4116--4128. Association for Computational Linguistics.

\bibitem[{Wang et~al.(2024)Wang, Ren et~al.}]{wang2024challenges}
Zihan Wang, Xiang Ren, and 1 others. 2024.
\newblock Llm multi-agent systems: Challenges and open problems.
\newblock \emph{arXiv preprint arXiv:2402.03578}.

\bibitem[{Wei et~al.(2022)Wei, Wang, Schuurmans, Bosma, Le et~al.}]{wei2022chain}
Jason Wei, Xuezhi Wang, Dale Schuurmans, Maarten Bosma, Harm Le, and 1 others. 2022.
\newblock Chain of thought prompting elicits reasoning in large language models.
\newblock In \emph{Proceedings of the 2022 EMNLP}, pages 2480--2493. Association for Computational Linguistics.

\bibitem[{Zhang et~al.(2025)Zhang, Zhang, Li et~al.}]{ma_survey_2025}
Ke~Zhang, Yali Zhang, Jinlong Li, and 1 others. 2025.
\newblock A comprehensive survey on multi-agent cooperative decision-making.
\newblock \emph{Information Fusion}.
\newblock ArXiv:2503.13415.

\bibitem[{Zhao et~al.(2024)Zhao, Khanh, and Daumé III}]{zhao-etal-2024-successfully}
Lingjun Zhao, Nguyen~X. Khanh, and Hal Daumé III. 2024.
\newblock \href {https://doi.org/10.18653/v1/2024.emnlp-main.42} {Successfully guiding humans with imperfect instructions by highlighting potential errors and suggesting corrections}.
\newblock In \emph{Proceedings of the 2024 Conference on Empirical Methods in Natural Language Processing}, pages 719--736.

\end{thebibliography}

\appendix
\section{Appendix} \label{sec:appendix}

%\subsection{Additional Analyses}
\subsection{Papers (SLR)-Wise Analysis}
\begin{figure}[H]
  \centering
  \includegraphics[width=\columnwidth]{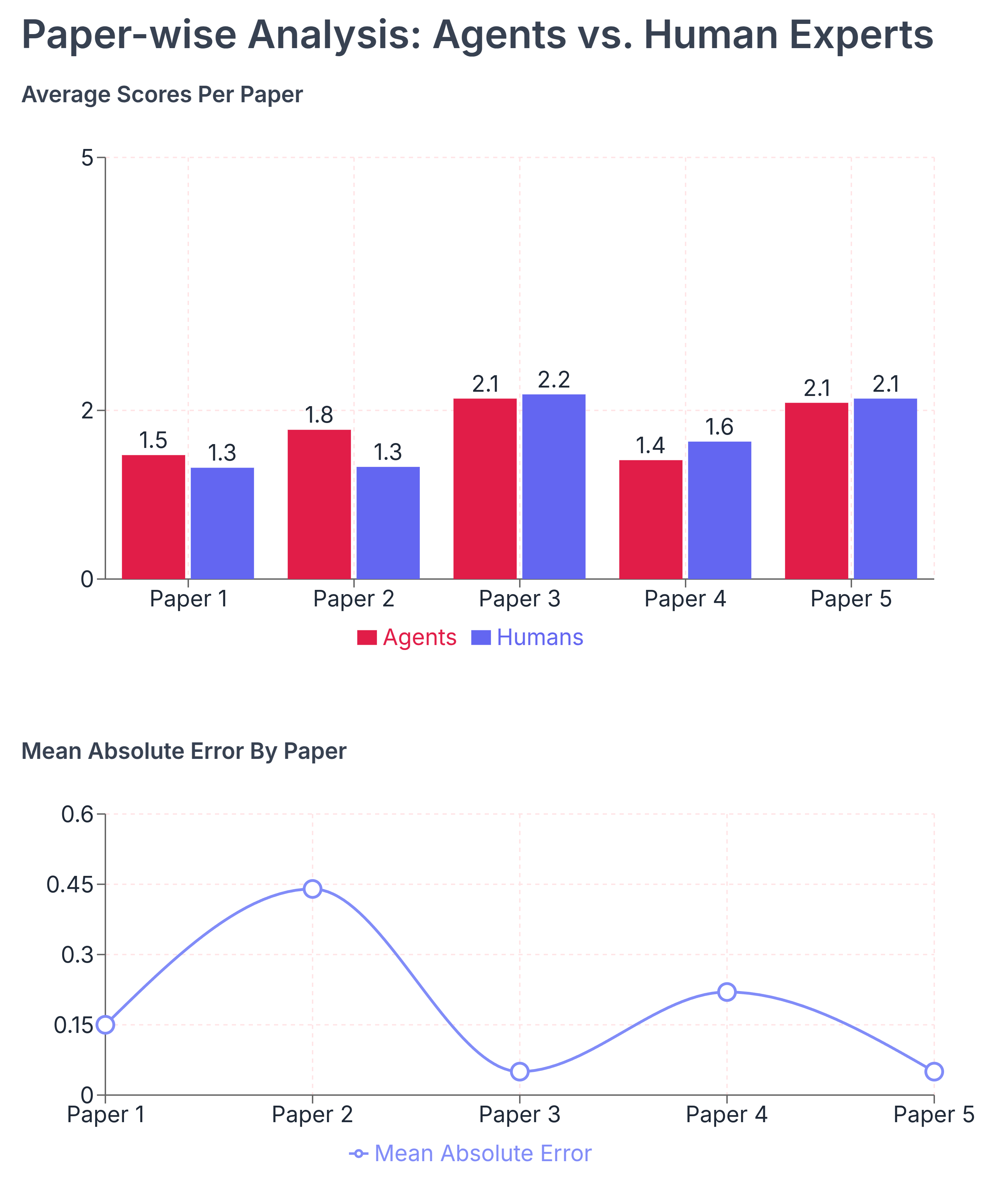}
  \caption{Paper‑wise comparison of average scores (Agents vs.\ Humans).}
  \label{fig:paper_analysis}
\end{figure}

Figure~\ref{fig:paper_analysis} presents a paper‑wise comparison of average PRISMA scores assigned by our MAS‑LLM system versus human experts across five SLRs. The mean absolute error (MAE) by paper ranges from 0.05 (Paper 3) to 0.44 (Paper 2), demonstrating consistently strong alignment and identifying specific instances for targeted model refinement.

\subsection{Human Experts' Scores}
\begin{figure}[H]
  \centering
  \includegraphics[width=\columnwidth]{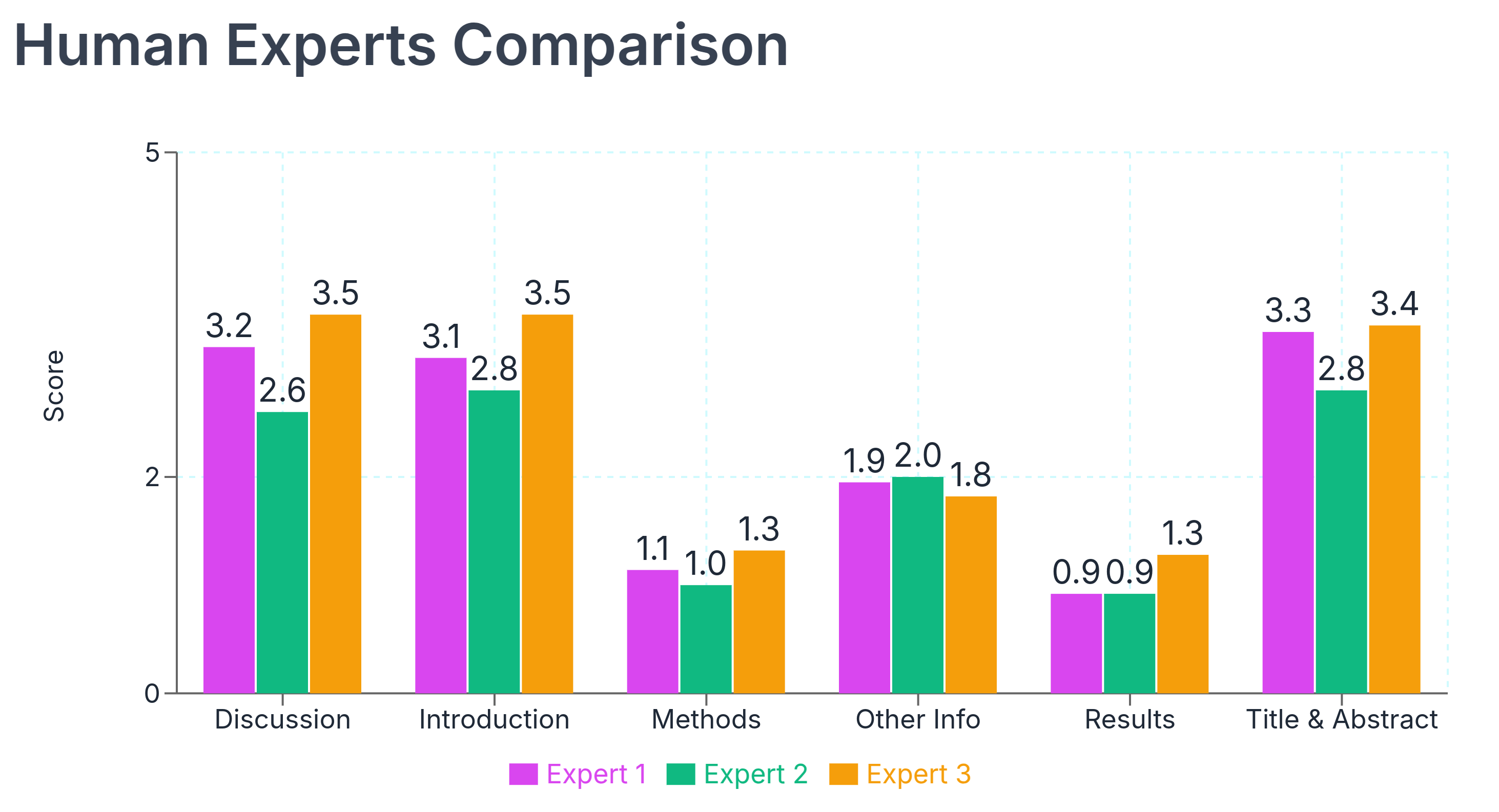}
  \caption{Variation in human expert scores by society.}
  \label{fig:expert_comparison}
\end{figure}

Figure~\ref{fig:expert_comparison} illustrates inter‑expert variability across PRISMA checklist societies. Methods sections exhibit the highest reviewer disagreement---reflecting the inherent complexity of methodological assessments---whereas Title \& Abstract sections achieve the greatest consensus.

\subsection{Inter-Expert Reliability Analysis}
Table~\ref{tab:human_agreement} summarizes inter‑expert reliability metrics, including Intraclass Correlation Coefficient (ICC), Krippendorff's Alpha, and average Pearson~$\rho$, all exceeding 0.88. These high values confirm the robustness of our expert benchmark and substantiate the validity of comparing agent outputs against human ratings.

\begin{table}[H]
\small
  \centering
  {%
    \renewcommand{\arraystretch}{1.3}%
    \begin{tabular}{|p{0.17\textwidth}|p{0.07\textwidth}|p{0.15
    \textwidth}|}
      \hline
      \textbf{Metric}                   & \textbf{Value} & \textbf{Interpretation}    \\ \hline
      Intraclass Correlation Coefficient (ICC)   & 0.924          & Excellent reliability      \\ \hline
      Krippendorff's Alpha              & 0.889          & Strong agreement           \\ \hline
      Avg.\ Inter‑Human Pearson~$\rho$  & 0.898          & Strong correlation         \\ \hline
    \end{tabular}
  }
  \caption{Inter‑expert agreement metrics.}
  \label{tab:human_agreement}
\end{table}

\end{document}